\documentclass[]{acmart}

\usepackage{algpseudocode, algorithm}
\AtBeginDocument{%
  \providecommand\BibTeX{{%
    \normalfont B\kern-0.5em{\scshape i\kern-0.25em b}\kern-0.8em\TeX}}}

\begin{document}

\title{An Interpretable Rule Creation Method for Black-Box Models based on Surrogate Trees - SRules}

\author{Mario Parrón Verdasco}
\author{Esteban García-Cuesta}
\affiliation{%
   \institution{Departamento de Inteligencia Artificial, Universidad Politécnica de Madrid}
   \city{Madrid}
   \country{Spain. }
   Corresponding author: esteban.garcia@upm.es}
\begin{abstract}
As artificial intelligence (AI) systems become increasingly integrated into critical decision-making processes, the need for transparent and interpretable models has become paramount. In this article we present a new ruleset creation method based on surrogate decision trees (SRules), designed to improve the interpretability of black-box machine learning models. SRules balances the accuracy, coverage, and interpretability of machine learning models by recursively creating surrogate interpretable decision tree models that approximate the decision boundaries of a complex model. We propose a systematic framework for generating concise and meaningful rules from these surrogate models, allowing stakeholders to understand and trust the AI system's decision-making process. Our approach not only provides interpretable rules, but also quantifies the confidence and coverage of these rules. The proposed model allows to adjust its parameters to counteract the lack of interpretability by precision and coverage by allowing a near perfect fit and high interpretability of some parts of the model . The results show that SRules improves on other state-of-the-art techniques and introduces the possibility of creating highly interpretable specific rules for specific sub-parts of the model.
\end{abstract}





\maketitle

\section{Introduction}
Although the origins of artificial intelligence (AI) can be traced back several decades, there is presently a widely held agreement that intelligent machines equipped with abilities to learn, reason and adapt hold paramount importance. Recently, the sophistication of AI-powered systems has advanced to a degree where the need for human intervention in their design and deployment has become almost negligible.

Over the past few years, non-interpretable techniques such as deep learning neural networks or ensemble methods, commonly referred to as black-box techniques \cite{castelvecchi2016can}, have emerged. Despite their lack of interpretability, these methods have demonstrated remarkable efficacy in solving intricate problems. 

Conversely in fields such as medicine~\cite{app131910778}, law, or defense, where decisions resulting from AI methodologies ultimately impact human lives, there is a growing need for comprehending how such decisions are generated. Explainable AI (XAI) aims to bridge the gap between the remarkable predictive capabilities of complex machine learning models and the human need for understanding and trust. \cite{gunning2017explainable,xIA}.
Based on the definition given by Barredo et al.\cite{xIA} \textit{" Given an audience, an explainable Artificial Intelligence is one that produces details or reasons to make its functioning clear or easy to understand."}


Our proposed method addresses the issues inherent in ensemble models like random forest \cite{RandomForest} or deep neural networks, although it can be extended to any black-box model. A black-box model can be formulated as $b : X \to y, b(x) = y$ with $b\in B$, where $B \subset H$ gives the model's hypothesis space~\cite{ali_explainable_2023}. For instance, $B =$ {random forest with 100 estimators}. Such models lose the ability to provide explanations for their results as a decision tree or a system of rules does.
Supervised ML is aimed at minimizing a given error metric (usually RMSE or hinge loss for binary classification)
\[h^* = arg\,min_{h\in H}\frac{1}{|n|}\sum_{x\in X}^n L(h(x_i),y_i)\],
where $h^*$ is the optimized model with the samllest loss. 

Thus, our proposition advocates for a post-hoc approach that commences with a black-box machine learning and subsequently generates a surrogate interpretable rule-based classifier based on the most important features obtained for that black-box.  Separating a black-box model from its explanation, according to~\cite{tulio2016model}, provides better accuracy, flexibility, and usability. Surrogate models may be classified as global or local and it can be supervised by fitting 

\[S^* = arg\,min_{w\in I}\frac{1}{|X|}\sum_{x\in X}^n FS(F(x),B(x))\]
being $w$ the hyperparameters of the surrogate model $S^*$, and $X=\{x_1,x_2,\ldots,x_n\}$ a subset of the entire training dataset $D$, or unsupervised assuming some inherent structure or patterns that can be explored and extracted from the black-box model to construct an interpretable model. 
In this paper we present an unsupervised and local method that focus on ensembles (e.g. random forest) and has inherent capabilities for obtaining feature importances (e.g. gini impurity o permutation test in random forest\cite{breiman2001random}) but other XAI post-hoc model agnostic techniques (e.g. SHAP\cite{SHAP} or LIME~\cite{tulio2016model}) can also be used to obtain the feature importance for other ML models that doesn't have an specific feature importance extraction method. 

There are other approaches that can obtain a global set of rules from random forest combining the individual set of rules somehow. Examples of these type of methods are RuleFit (RF+RFIT) \cite{RuleFit} and RuleCosi (RF+RC) \cite{RuleCosi}. The RuleFit algorithm consists of using both original features and new decision rules generated from decision trees. These decision rules capture interactions between the original features. RuleFit transforms each path through a tree into a decision rule by combining split decisions, and discards node predictions, using only the splits in the decision rules. The resulting model is a sparse linear model that captures complex interactions between the original features and the decision rules. In a manner akin, RuleCosi \cite{RuleCosi} leverages each decision tree path to derive a set of rules. These rules are subsequently combined and pruned, keeping only the combined rules that have better results, to yield a more parsimonious set of rules. In this paper we compare our results with RuleCosi and RuleFit to show our increased performance and more importantly the ability of the method to learn high accuracy rule sets for some parts of the analyzed problem.


\section{Rule-set Creation Based on Conditional Information Gain}
A single decision tree can be converted into a set of production rules. The number of production rules that can be extracted from a decision tree is the same as the number of leaf nodes of the tree. In~\cite{RuleCosi,RuleCosi_second,SirikulviriyaIntegrationOR,benard2021,RuleFit,inTrees} different authors combine, using different methods, the set of rules obtained from the different weak learners to build a set of rules that describe the different parts of the problem that each of these weak learners has learned. Instead, we create recursively a set of rules based on surrogate tree models created using the most important features of an ensemble learned recursively using only the set of data that has not been already covered by the learned set of rules.  The most important features are obtained by the analyzing the impurity associated to the application of a feature to a node and averaging it over all the weak learners. Gini impurity~\cite{breiman2001random} measures the decrease on impurity when a split is performed. A feature that contributes to reduce the impurity is considered important and vice-versa. Gini impurity is defined as: $\hat{\Gamma}(t) = \sum_{j=1}^{J}\hat{\Phi}(t)(1-\hat{\Phi}(t)$, where $\hat{\Phi}(t)$ is the class frequency for class $j$ in the node $t$. The decrease of impurity is the difference between a node's impurity and the weighted sum of the impurity measures of the two child nodes (the Gini index). 

In the next, the definition of a production rule is presented, and the performance measures of a
set of production rules are explained following the same nomenclature as~\cite{RuleCosi}.

\textbf{Definition 1} (Output of a boosting ensemble of binary decision trees). The boosting ensemble of the binary decision trees produces a classification model $H(x) = \{(h_n(x), w_n)\}$ for $n = 1$ to $N$ week learners. A base decision tree $h_n(x)$ was derived from data examples in the training set D, and w contains weights $w_n$ associated to each decision tree $h_n(x)$.

\textbf{Definition 2} (Production rule). Let a rule r be in the form r:
$A\to y$, where A is a set of conditions $a \in A$, also called left hand side of the rule (LHS), and $y$ is the consequent class, also called the right hand side of the rule (RHS). A condition $a$ is an evaluation expression between an attribute $x_{ij}$ and a value using an operator such as equal ($=$), not equal ($\ne$), less than ($<$), greater than ($>$), less than or equal to ($le$), or greater than or equal to ($\ge$). In our case we are limiting the algorithm to equal ($=$), and not equal ($\ne$) operators (binary inputs).

Production rules are grouped into rulesets $R$. A rule $r$ covers a record $x_i$ if the conditions of the LHS of a rule are satisfied using the attributes of $x_i$. To measure the performance of a ruleset, each record of a dataset $D$ is evaluated to check if it triggers at least one rule $r$ of the ruleset $R$. Then, the coverage $c$ of a ruleset $R$ over a dataset $D$ is given by the fraction of records in $D$ that trigger, at least, one rule of the ruleset $R$. Aditionally, the accuracy or confidence factor of the rule is given by the fraction of records in $D$ triggered by, at least, one rule of the ruleset $R$ whose class labels are equal to $y$.

For measuring the performance of the ensemble models and the SRules model, the confusion matrix is utilized. 

\textbf{Definition 3} (Rule generation from a surrogate decision tree extracted from ensemble).
Given an ensemble of weak learners $H(X)$ (e.g. decision trees) which contains multiple weak learners $h_n(x)$ for $n=1$ to $N$ and a weight vector $w$ of size $N$, the problem of rule generation is to derive a set of simplified production rules $R*$ from $H(X)$. The ruleset $R^*$ is required to maintain or increase the performance of $H(X)$ with a minimum set of simplified production rules $R^*$ and the maximum possible coverage $c$.  The first step of the methodology consists in taking the output of a trained ensemble and extracting an ordered set of features importance $F^*$ of the model. Using $F^*$, a surrogate binary tree ($SBT$) is constructed starting the root node with the most important feature and splitting each node with the next most important features. On the second step, we use a conditional $\tilde{\chi}^2$ importance~\cite{Wang2010} pattern selection method to select those branches of the binary tree that are relevant to the problem under the null hyptothesis that no pattern exist for the splitting of a node $j$, conditional to the previously seen nodes of the branch. 

Let $L_i^j$ be the list of ordered features $F^*$ that precede feature $f_i$ in its split of the $j-th$ node of the tree. Let be the pattern under analysis the set $S_i=\{X_i^j| L_i^j, j=1,\ldots,n_i\}$ and the pattern $P_i^j = \{X_i^j, L_i^{j},\ldots,L_i^{j-n}\}$ importance is then obtained using the data samples assigned to the tree by: 

\[P_i^j=\sum_{k=1}^{c}\sum_{s=1}^{ns} \frac{(O_{ks} - E_{ks})^2}{E_{ks}}\]

being $c$ the number of categories of the feature $i$, and $s$ the number of splits of the node $L_i^j$, and

\[E_{ks}=T_{ks} * Freq_{ki}\]

being $Freq_{ki}$ the frequency of category $k$ for node-feature $i$ (parent) and $T_{ks}$ the total number of samples of node-feature $s$ (child). This has been also studied by (in an unpublished work) to extract feature importance in an heuristic way as the difference in expectacions in a decision path knowing that the estimated expectation of the model output conditioned on the set $F^*$ of feature values is $f_x(F^*) \approx E[f(x)|x_{F^*}]$\cite{saabas}. Last, we use these empirical values to obtain the p-value associated to the $P_i^j$ importance value. This p-value controls the certainty that the algorithm SRules has over the pattern $P_i^j$ under analysis and it has to be tuned as a parameter of the algorithm base on the certainty needed for a specific application or the pursued level of interpretation. 

After the statistical analysis of every possible pattern is obtained the generated patterns are translated into rules and ordered by accuracy and simplicity (rule length). We detect duplicates (p.ej. rules that contain another rules) and simplify them by quality of interpretation capabilities ~\cite{xIA} (the more general pattern is stored). The process is repeated until all the ruleset are processed. The resulting ruleset $R^{1*}$ for iteration 1 is then stored and a new iteration starts using those samples from the dataset that are not covered by $R^{1*}$. The final ruleset $R^*$ is then obtained by a combination of the rulesets obtained for each iteration $R*=\{R^{1*},R^{2*},\ldots,R^{i*}\}$.

\subsection{SRules Algorithm}
SRules is a rule extraction algorithm that given a black-box machine learning model greedily searches for the most important features and the associated tree structures to find a subset of rules that provides interpretation over the dataset. A description of the steps taken in the proposed algorithm, SRules, is presented in \ref{alg:SRules}. The input of the algorithm is the output of the black-box model (in our case is an ensemble $H(X) = \{h_n(x), w_n\}$ described in definition 1. The first phase of the algorithm is the extraction of the feature importance of the given model H(X). We use  the permutation test technique described in~\cite{breiman2001random} and Gini criteria to obtain the feature importance values for the tested ensemble methods (random forest, adaboost, catboost). A minimum value of importance is needed to set a maximum level of depth of the binary tree that we construct, otherwise for datasets that have a very large number of features the construction of the tree would be costly. 

\begin{algorithm}
\caption{\textbf{SRules}: SRules algorithm for creating rulesets from black-box models using feature importance FI}
\label{alg:SRules}
\begin{algorithmic}
\State \textbf{input}: black-box ensemble model \textbf{H}, \textbf{minImp} minimum feature importance value in order to be extracted, \textbf{minInsNode} (the minimum number of instances that each node must have to be included as a rule), \textbf{$\alpha$} (the minimum precision that the rule must have in order to be included as a rule; assures high precision), \textbf{p-significance} (the p-value used for selecting or not a branch-rule $L_i^j$)  
\State \textbf{output:} $R^*$ the minimum set of interpretable rules
 \State  

\State FI $\leftarrow$ ExtractFeatureImportance(H, minImp)
\State R = $\emptyset$; $P^*=\emptyset$
\State SBT = init($FI_1$)
\State SBT $\leftarrow$ BinaryTreeGenerator(SBT, FI, minInsNode, $FI_2$, $FI_1$)
    \For{each $L_i^j$ \textbf{in} SBT} \Comment{We walk through the tree branches}
        \State $ns_i$ $\leftarrow$ GetChildren(i)
        \State $p_i^j$ $\leftarrow$ Test$\tilde{\chi}^2$($ns_i$) \Comment{obtain the p-value of the pattern $P_i^j$}
        \State \If{$p_i^j < $ p-significance}    			
            \State $P^*$ $\leftarrow$ append($L_i^j$) \Comment{the conditional pattern $L_i^j$ is added to the set of rules patterns $P^*$}
        \EndIf
   \EndFor
\State R $\leftarrow$ RuleCategorization($P^*$, $\alpha$)
\State $R^*$ $\leftarrow$ Prune(R)
\State
\State \Return $R^*$

\end{algorithmic}
\end{algorithm}

The SBT is initialize with the most important feature and the the binary tree algorithm is called to construct the SBT structure. The algorithm is detailed at (\ref{alg:binary_tree_generator}) and it uses the minInstances value to control if a new node is created or not (a minimum value of 5 is set for statistical reasons).  Once the tree is built then we check if the combination set of interpretable branches (of different possible sizes) are a pattern of interest. For that purpose we use the conditional $\tilde{\chi}^2$ test that empirically can be computed as $P_i^j$ as described earlier. The selection or not of that particular pattern $P_i^j$ is decided using the statistical significance threshold of the pattern for classifying the instances that accomplish with that pattern (i.e. they share the same values for the pattern list $L_i^j$). In that case the pattern is added to the set of patterns $P^*$ associated to the dataset. After obtaining the set of patterns $P^*$, it is necessary to define on which class they will be assigned to convert them in a set of rules $R$. For this purpose, the class with the highest precision of the training set will be assigned, and if it does not exceed $minPrecision$, it will not be included as a rule and the pattern will be discarded. Likewise, the rules will be ordered from the highest to the lowest level of precision, where the most precise are the first and the least precise will be the last.

\begin{algorithm}
\caption{\textbf{BinaryTreeGenerator}: generates a surrogate binary tree (SBT) based on feature importance FI}
\label{alg:binary_tree_generator}
\begin{algorithmic}
\State \textbf{input}: \textbf{SBT}, \textbf{FI} ordered list of feature importance, \textbf{minInsNode}, \textbf{currentNode}, \textbf{parentNode})
\State \textbf{output}:
\State newNode  $\leftarrow$  CreateNode(currentNode, parentNode)
\If{ numInstances(newNode) $>=$ minInsNode} \Comment{Compare instances of the new node}
  \State SBT $\leftarrow$ add newNode		
  \State currentNode $\leftarrow$ nextFeature(FI) 		
  \For{child \textbf{in} currentNode}
       \State BinaryTreeGenerator(SBT, FI, minInsNode, child, newNode)   \Comment{New parentNode is the current node}			
    \EndFor
  \State  return SBT
\EndIf
\end{algorithmic}
\end{algorithm}

There are some rules that may be duplicate, i.e. one more general rule may include others more specific. In such cases, the more general rules have priority (they are more interpretable and contain a larger number of instances which represents more knowledge) and the others will be discarded reducing also the number of rules and easy the explanation. To do this, we iterate through the set rules $R*$ from lower priority to higher priority and check if there is any higher priority rule that contains it. If so, the rules of lower priority that are contained in those of higher priority will be eliminated from the final set $R^*$.

\subsection{Some thoughts over the method and extension}
It can be observed that SRules only iterates once over the entire dataset and then obtains a set of interpretable rules based on a surrogate decision tree  model. SRules uses a minimum precision threshold $\alpha$ during the categorization step of pattern discovery to guarantee minimum performance. If $\alpha$ is high (i.e., we require high precision on the analyzed pattern $L_i^j$) a smaller set of patterns, and also associated rules, will be obtained, but they will represent with high confidence the instances that are included in that pattern. On the other hand, only a subset of instances will be represented in those rules and, thereof, the coverage will be smaller (we will discuss more about this in the results section). At this point it is important to note that our final system does not have a default rule for any of the instances that are not covered by any rule. We could define a default rule for those instances using the expected value but we focus on obtaining interpretable rules as close to reality as possible and,  therefore, if the performance of the classifier is bad in some instances we  prefer not to explain. 

Instead we propose an extension of the method SRules to address this problem recursively (Recursive SRules). We adjust the $\alpha$ value to obtain representative enough rules according to the perfomance criteria (e.g. F1 classification) and then we obtain the ruleset $R_1$ associated to that criteria using SRules. Afterwards, we assign instances $x \in X$ to the ruleset $R_1$ if they fulfill the LHS conditions obtaining a new set of covered instances $X'$ and another set of not-covered instances $\overline{ X}': X' \cup \overline{X}' = X$. Then we apply again SRules with $\overline{X}$ obtaining new set of rules for a second iteration $R_2$; for $n$ iterations we will obtain $\{R_1,R_2,\ldots,R_n\}$ and a final set $R^*$ is then obtained by applying the prune function to those rules $R^* = Prune(\{R_1, R_2,\ldots,R_n\})$. We call to this second approach Recursive SRules algorithm (RSRules). 
RSRules gives some flexibility to the algorithm and allows for greater coverage while maintaining efficiency, but penalizing explainability by obtaining a greater number of rules (in the best case will be a similar number).
\section{Experiments and Results}\label{heads}

\subsection{Experiments design}
In order to prove the effectiveness of the proposed method, several experiments were conducted. In this section, the context of these experiments is presented. 

\subsubsection{Datasets}
Five public datasets available in the UCI Machine Learning Repository~\cite{UCI2013} have been used (divorce, kr-vs-kp, wisconsin, SPECT, tic-tac-toe) and additionally a survival Covid-19 dataset (salud-covid)\cite{covid19dataset} has been included. Given that the current state of the SRules algorithm only supports binary attributes, a one hot encoding \cite{Stevens1946} has been applied to create binary attributes. The datasets have been selected with different ranges of attributes and instances to demonstrate the effectiveness of the method on different contexts. The datasets are well-known real world cases commonly used as benchmark in the machine learning field. The properties of these datasets can be seen in Table \ref{table:datasets}.  It details the number of instances, attributes, binary attributes, categorical attributes as well as the total number of attributes after pre-processing.

\begin{table}
    \centering
    \caption{Summary of the properties of the 5 datasets used in the experiments}
\resizebox{0.9\columnwidth}{!}{%
    \begin{tabular}{|l|c|c|c|c|c|c|c|}
    \hline
        \textbf{Dataset} & \textbf{Class  (minor; major)} & \textbf{\# inst.} & \textbf{\# atts.} & \textbf{\# atts. bit.} & \textbf{\# atts. cat.} & \textbf{\# atts. tot.} \\ \hline
        divorce & (yes; no) & 170 & 54 & ~ & 54 & 270 \\ 
        kr-vs-kp & (win; no win) & 3196 & 36 & ~ & 36 & 73 \\
        wisconsin & (malignant; benign) & 683 & 9 & ~ & 9 & 90 \\ 
        SPECT & (positive; negative) & 267 & 22 & 22 & ~ & 22 \\
        salud-covid & (surv.; not surv.) & 1744 & 119 & 119 & ~ & 119 \\
        tic-tac-toe & (positive; negative) & 958 & 9 & ~ & 9 & 27 \\  \hline       
    \end{tabular}
    }
    
    \label{table:datasets}
\end{table}

\subsection{Algorithms and experiment validation}
Four ensemble based algorithms, random forest (RF) \cite{RandomForest}, Cat Boost Classifier (CAT) \cite{10.5555/3327757.3327770}, Ada Boost Classifier (ADA) \cite{FREUND1997119}, Gradient Boost Classifier (GBC) \cite{FRIEDMAN2002367,10.1214/aos/1013203451}, were chosen to prove the applicability and performance of SRules. The purpose is to cover different approaches to deal with class imbalance problem. Moreover, RuleFit (RF+RFIT)  \cite{RuleFit} and RuleCosi (RF+RC) \cite{RuleCosi} were also used for comparative purposes given that are state of the art methods that also creates interpretable rules from ensemble models. We also included in our experiment design the most simple interpretable algorithm, decision trees (DT), as a baseline to compare with. Following the experimentation line of proposed in the RuleCOSI+ paper \cite{RuleCosi,RuleCosi_second}. A stratified cross validation with 10-fold repeated 3 times was performed. Within each fold, the hyperparameters of the ensemble are optimized in a grid-seach with cross validation of a 5-fold, once the ensemble method is trained the best ensemble will be applied to each rule extraction method for their training. Ensemble hyper-parameters vary the depth of the trees by $\{2,3,4,5,6\}$, and the number of estimators by $\{10, 25, 50, 100, 250, 500\}$. Likewise for RuleFit the recommended parameters have been taken, $\alpha = 30$, $include\_linear = True$. Similarly, the recommended parameters for RuleCosi+ have been taken,  $\alpha = 0.5$, $\beta = 0$, $c = 25\%$. For SRules, different parameters have been taken: $minImp \in \{ 0.05, 0.1\}$,  $\alpha \in \{ 0.9, 0.95\}$, $minInsNode \in \{5,7,10, 25, 50\}$, $p$-$significance \in \{ 0.95, 0.97\}$. The metric to be optimized by SRules is defined by Coverage x F1-score, the higher coverage and F1-score, the better. Note that we can limit the number of rules created using $minImp$, $minInsNode$, and $p$-$significance$ parameters. A large $minImp$ value will only take into account a few features that will be used during the construction of the SBT and thereof reduces the possible number of final rules $R^*$. $minInsNode$ parameter has a similar objective by establishing a minimum number of instances that each node of a rule must have, thereof a large value will also reduce the number of rules and its complexity (the length of the components of the rule). Finally, the $p$-$significance$ parameter also controls whether an SBT conditional pattern is selected. A high value will prevent the selection of that pattern and will also reduce the number of rules. In summary, these parameters allows to reduce the number of rules, increasing interpretability, but also reduce the number of instances covered by those rules (note that there is a trade-off between ease of of interpretation and coverage) and the optimization metric maximizes the F1-score and coverage constrained by those parameters' values. We used a stratified cross-validation with 10-fold repeated 3 times for each combination of ensemble model+rule generation method. We measure the coverage (percentage of instances that are covered by the generated rules $R^*$), the F1-score, and the number of rules (rl. num). The results are shown as the average of the stratified cross validation with 10-fold repeated 3 times, as well as its standard deviation.


\subsection{Experimental study}
In this section, we present the empirical results of the experiments using SRules and RSRules. First, to confirm that the method simplifies the result of the different selected ensemble algorithms. This is done by comparing the number of rules obtained with SRules with those obtained by the other state of the art algorithms (RuleCOSI and RuleFIT) for the different datasets and ensemble methods. Secondly, to evaluate whether the classification performance of the final ruleset is at least maintained compared to that of the original ensemble from which it is derived.

\begin{table}
    \centering
    \caption{Non Recursive SRules (SR). Black-box model average performance results.}

\resizebox{\columnwidth}{!}{%
    \begin{tabular}{c||c||c|c|cc|cc|cc}
    \hline
       \textbf{Method}  & \textbf{Coverage} & \textbf{DT F1} & \textbf{Ensemble F1} & \textbf{RFIT F1}& \textbf{RFIT rl. num.} & \textbf{RC F1}& \textbf{RC rl. num.} & \textbf{SR F1} & \textbf{SR rl. num.} \\
        Random Forest & 63.89±22.7 & 90.38±18.56 & 88.29±23.9 & 91.59±18.3 & 21.56±3.73 & 90.79±19.59 & 18.41±8.09 & 91.52±17.94 & 9.05±6.58 \\ 
        Gradient Boost & 61.27±23.74 & 86.72±25.16 & 89.22±24.16 & 88.63±23.93 & 21.49±3.31 & 88.51±24.66 & 17.28±13.07 & 88.65±23.74 & 8.92±7.46 \\ 
        Ada Boost & 56.57±26.24 & 83.91±29.74 & 85.88±29.63 & 84.17±30.82 & 21.15±3.38 & 68.64±36.05 & 2.81±1.73 & 81.84±35.02 & 19.39±25.45 \\ 
        Cat Boost & 58.96±24.03 & 91.62±18.16 & 65.84±29.46 & 93.15±17.42 & 21.24±3.25 & 79.64±32.71 & 3.21±1.91 & 93.01±17.5 & 5.91±3.98 \\ \hline
        \textbf{Global Average} & 60.17±24.31 & 88.16±23.56 & 82.31±28.53 & 89.38±23.45 & 21.36±3.42 & 81.89±30.21 & 10.43±10.76 & 88.76±24.91 & 10.82±14.69 \\ \hline
    \end{tabular}
    }    
    \label{results_table_no_recursive_global}
\end{table}

\begin{table}
    \centering
    \caption{Non Recursive SRules (SR). Black-box model two-sample t-test analysis for equal means results (SRules vs. model p-value)}

\resizebox{\columnwidth}{!}{%
    \begin{tabular}{c||c||c|c|cc|cc}
    \hline
       \textbf{Method}  & \textbf{Coverage}& \textbf{DT p-value F1} & \textbf{Ensemble p-value F1}& \textbf{RFIT F1 p-value} & \textbf{RFIT rl. num. p-value} & \textbf{RC F1 p-value} & \textbf{RC rl. num. p-value} \\
        Random Forest & 63.89±22.7 & 0.5537 & 0.1471 & 0.9731 & <0.0001 & 0.7101 & <0.0001 \\ 
        Gradient Boost & 61.27±23.74 & 0.4554 & 0.8202 & 0.9939 & <0.0001 & 0.9558 & <0.0001 \\ 
        Ada Boost & 56.57±26.24 & 0.5465 & 0.2387 & 0.5042 & 0.3579 & 0.0005 & <0.0001 \\ 
        Cat Boost & 58.96±24.03 & 0.4601 & <0.0001 & 0.9380 & <0.0001 & <0.0001 & <0.0001 \\ \hline
        \textbf{Global Average} & 60.17±24.31 & 0.6400 & <0.0001 & 0.6222 & <0.0001 & <0.0001 & 0.5640 \\ \hline
    \end{tabular}
    }    
    \label{results_table_no_recursive_global_pvalue}
\end{table}

\begin{table}
    \centering
    \caption{Recursive SRules (RSR) -- Black-box model average performance results.}

\resizebox{\columnwidth}{!}{%
    \begin{tabular}{c||c||c|c|cc|cc|cc}
    \hline
        \textbf{Method}  & \textbf{Coverage} & \textbf{DT F1} & \textbf{Ensemble F1} & \textbf{RFIT F1} & \textbf{RFIT rl. num.} & \textbf{RC F1} & \textbf{RC rl. num.} & \textbf{RSR F1} & \textbf{RSR rl. num.} \\ \hline
        Random Forest & 73.18±21.63 & 89.35±18.78 & 86.75±24.73 & 91.25±17.4 & 21.41±3.69 & 90.98±17.93 & 18.36±8.52 & 91.44±17.21 & 13.94±11.39 \\ 
        Gradient Boost & 73.51±24.37 & 88.4±19.77 & 91.6±18.36 & 90.51±18.64 & 21.48±3.41 & 90.29±19.54 & 17.18±13.52 & 89.43±18.27 & 13.71±11.01 \\ 
        Ada Boost & 69.92±23.53 & 88.73±17.94 & 91.47±17.65 & 90.21±19.64 & 21.48±3.35 & 72.31±31.04 & 2.74±1.64 & 87.94±22.06 & 28.06±34.71 \\ 
        Cat Boost & 73.64±19.24 & 91.46±13.45 & 69.27±25.88 & 93.48±11.54 & 21.57±3.87 & 79.22±32.29 & 3.25±2.02 & 93.29±11.14 & 11.02±6.94 \\ \hline
        \textbf{Global Average} & 72.57±22.29 & 89.48±17.66 & 84.77±23.76 & 91.36±17.11 & 21.48±3.58 & 83.2±27.13 & 10.38±10.96 & 90.52±17.69 & 16.68±20.47 \\  \hline 
\end{tabular}
    }    
    \label{results_table_recursive_global}
\end{table}

\begin{table}
    \centering
    \caption{Recursive SRules (RSR). Black-box model two-sample t-test analysis for equal means results (RSRules vs. model p-value).}

\resizebox{\columnwidth}{!}{%
    \begin{tabular}{c||c||c|c|cc|cc}
    \hline
       \textbf{Method}  & \textbf{Coverage}& \textbf{DT p-value F1} & \textbf{Ensemble F1 p-value}& \textbf{RFIT F1 p-value} & \textbf{RFIT  rl. num. p-value} & \textbf{RC F1 p-value} & \textbf{RC rl. num. p-value} \\
        Random Forest & 73.18±21.63 & 0.2708 & 0.0373 & 0.9170 & <0.0001 & 0.8044 & <0.0001 \\ 
        Gradient Boost & 73.51±24.37 & 0.6104 & 0.2606 & 0.5779 & <0.0001 & 0.6666 & 0.0078 \\ 
        Ada Boost & 69.92±23.53 & 0.7090 & 0.0945 & 0.3032 & 0.0117 & <0.0001 & <0.0001 \\ 
        Cat Boost & 73.64±19.24 & 0.1610 & <0.0001 & 0.8750 & <0.0001 & <0.0001 & <0.0001 \\ \hline
        \textbf{Global Average} & 72.57±22.29 & 0.2651 & <0.0001 & 0.3611 & <0.0001 & <0.0001 & <0.0001 \\ \hline
    \end{tabular}
    }    
    \label{results_table_recursive_global_pvalue}
\end{table}

The measure of interpretability used in this work is the number of rules of the classification model (i.e. ruleset $R^*$).  Table~\ref{results_table_no_recursive_global} shows a summary of the metrics: coverage, F1-score (F1), and number of rules; for the different ensemble methods and decision tree, for the non-recursive SRules approach and the methods RuleFIT and RuleCOSI. Table~\ref{results_table_recursive_global} shows the same information but for the recursive version of SRules. A complete list of results over the different datasets can be seen at tables~\ref{table:results_table_non_recursive} and~\ref{table:results_table_recursive}.

\begin{table}
 \centering
 \caption{Non Recursive SRules (SR). Comparison p-value performance per model and dataset.}

\resizebox{\columnwidth}{!}{%
 \begin{tabular}{c||c||c||c|c|cc|cc}
 \hline
       \textbf{Method} & \textbf{Dataset} & \textbf{Coverage}& \textbf{DT F1 p-value} & \textbf{Ensemble F1 p-value}& \textbf{RFIT F1 p-value} & \textbf{RFIT rl. num. p-value} & \textbf{RC  F1 p-value} & \textbf{RC  rl. num. p-value}  \\ \hline 

 RF & divorce & 92.55±11.02 & 0.7951 & 0.7732 & 0.844 & <0.0001 & 0.9849 & <0.0001 \\ 
 RF & kr-vs-kp & 82.46±4.33 & 0.251 & 0.2769 & 0.8283 & <0.0001 & 0.1206 & 0.9062 \\ 
 RF & wisconsin & 73.71±10.36 & 0.6767 & 0.0282 & 0.0648 & <0.0001 & 0.1525 & <0.0001 \\ 
 RF & SPECT & 34.73±15.18 & 0.9706 & 1.0000 & 0.9836 & <0.0001 & 0.9652 & <0.0001 \\ 
 RF & salud-covid & 54.13±6.53 & 0.3703 & 0.0004 & 0.7395 & <0.0001 & 0.2599 & <0.0001 \\ 
 RF & tic-tac-toe & 45.79±6.04 & 0.0625 & 0.6905 & 0.2039 & <0.0001 & 0.248 & <0.0001 \\ \hline
 RF & Total Average & 63.89±22.7 & 0.5537 & 0.1471 & 0.9731 & <0.0001 & 0.7101 & <0.0001 \\ \hline
 GBC & divorce & 89.8±15.52 & 0.8773 & 0.8184 & 0.8184 & <0.0001 & 0.8158 & <0.0001 \\ 
 GBC & kr-vs-kp & 76.39±6.74 & <0.0001 & <0.0001 & 0.4208 & <0.0001 & 0.0037 & <0.0001 \\ 
 GBC & wisconsin & 70.47±12.32 & 0.5055 & 0.0883 & 0.0311 & <0.0001 & 0.2064 & <0.0001 \\ 
 GBC & SPECT & 29.82±20.71 & 0.981 & 1.0000 & 0.9892 & <0.0001 & 0.9962 & <0.0001 \\ 
 GBC & salud-covid & 55.16±7.27 & 0.0242 & 0.7308 & 0.4645 & <0.0001 & 0.1752 & <0.0001 \\ 
 GBC & tic-tac-toe & 45.97±9.39 & 0.0015 & 0.0003 & 0.1929 & 0.7745 & 0.0051 & 0.5704 \\  \hline
 GBC & Total Average & 61.27±23.74 & 0.4554 & 0.8202 & 0.9939 & <0.0001 & 0.9558 & <0.0001 \\  \hline

 ADA & divorce & 90.98±16.33 & 0.8868 & 0.8548 & 0.8574 & <0.0001 & 0.9582 & 0.0236 \\ 
 ADA & kr-vs-kp & 62.01±12.06 & <0.0001 & 0.2354 & 0.0065 & <0.0001 & <0.0001 & <0.0001 \\ 
 ADA & wisconsin & 75.75±6.43 & 0.8215 & 0.0194 & 0.0002 & <0.0001 & 0.0294 & 0.0001 \\ 
 ADA & SPECT & 17.89±11.32 & 0.8963 & 0.9788 & 0.9556 & <0.0001 & 1.0000 & 0.0039 \\ 
 ADA & salud-covid & 38.72±9.32 & 0.0276 & 0.0123 & 0.0747 & <0.0001 & 0.1239 & <0.0001 \\ 
 ADA & tic-tac-toe & 54.08±7.91 & <0.0001 & 0.0540 & 0.0019 & 0.0338 & <0.0001 & <0.0001 \\ \hline
 ADA & Total Average & 56.57±26.24 & 0.5465 & 0.2387 & 0.5042 & 0.3579 & 0.0005 & <0.0001 \\ \hline

 CAT & divorce & 91.18±14.39 & 0.6923 & <0.0001 & 0.9078 & <0.0001 & 0.8473 & <0.0001 \\ 
 CAT & kr-vs-kp & 49.27±11.0 & 1.0000 & <0.0001 & 0.9695 & <0.0001 & <0.0001 & 0.0364 \\ 
 CAT & wisconsin & 84.23±4.53 & 0.1186 & <0.0001 & 0.7844 & <0.0001 & 0.0005 & <0.0001 \\ 
 CAT & SPECT & 33.8±18.6 & 0.9298 & 1.0000 & 0.9858 & <0.0001 & 1.0000 & 0.1687 \\ 
 CAT & salud-covid & 46.57±8.81 & 0.2667 & <0.0001 & 0.9752 & <0.0001 & <0.0001 & <0.0001 \\ 
 CAT & tic-tac-toe & 48.71±6.08 & 0.0083 & <0.0001 & 0.5797 & <0.0001 & <0.0001 & <0.0001 \\  \hline
 CAT & Total Average & 58.96±24.03 & 0.4601 & <0.0001 & 0.9380 & <0.0001 & <0.0001 & <0.0001 \\  \hline

        \textbf{Global Average} & Total Average & 60.17±24.31 & 0.6400 & <0.0001 & 0.6222 & <0.0001 & <0.0001 & 0.5640 \\ \hline

 \end{tabular}
 } 
 \label{table:results_table_non_recursive_pvalue}
\end{table}

\begin{table}
 \centering
 \caption{Non Recursive SRules (SR). Average performance per model and dataset.}

\resizebox{\columnwidth}{!}{%
 \begin{tabular}{c||c||c||c|c|cc|cc|cc}
 \hline
\textbf{Method}& \textbf{Dataset}  & \textbf{Coverage} & \textbf{DT F1} & \textbf{Ensemble F1} & \textbf{RFIT F1} & \textbf{RFIT rl. num.} & \textbf{RC F1} & \textbf{RC rl. num.} & \textbf{SR F1} & \textbf{SR rl. num.} \\ \hline 
 RF & divorce & 92.55±11.02 & 91.75±18.0 & 94.33±18.08 & 93.9±18.12 & 26.47±3.53 & 93.06±18.14 & 5.7±1.18 & 92.97±18.15 & 2.9±0.66 \\ 
 RF & kr-vs-kp & 82.46±4.33 & 98.67±1.36 & 98.7±1.35 & 99.0±1.24 & 22.07±3.12 & 99.48±0.67 & 18.13±3.59 & 99.07±1.26 & 18.23±2.92 \\ 
 RF & wisconsin & 73.71±10.36 & 97.85±1.93 & 98.71±1.48 & 98.53±1.47 & 19±2.48 & 98.34±1.61 & 16.4±3.38 & 97.62±2.2 & 6.2±1.06 \\ 
 RF & SPECT & 34.73±15.18 & 89.02±24.83 & 89.25±24.88 & 89.39±24.9 & 22.23±2.34 & 89.53±24.85 & 19.6±4.59 & 89.25±24.88 & 1.43±0.68 \\ 
 RF & salud-covid & 54.13±6.53 & 65.99±18.77 & 48.95±23.94 & 68.8±19.7 & 18.03±1.61 & 64.39±21.62 & 25.2±8.08 & 70.52±20.04 & 10.17±2.34 \\ 
 RF & tic-tac-toe & 45.79±6.04 & 99.01±1.84 & 99.78±0.7 & 99.91±0.34 & 21.53±2.01 & 99.91±0.51 & 25.4±4.54 & 99.71±0.8 & 15.37±3.73 \\  \hline
 RF & Total Average & 63.89±22.7 & 90.38±18.56 & 88.29±23.9 & 91.59±18.3 & 21.56±3.73 & 90.79±19.59 & 18.41±8.09 & 91.52±17.94 & 9.05±6.58 \\  \hline

 GBC & divorce & 89.8±15.52 & 88.12±30.01 & 88.71±30.19 & 88.71±30.19 & 25.03±3.43 & 88.73±30.19 & 5.43±0.73 & 86.92±29.75 & 2.7±0.7 \\ 
 GBC & kr-vs-kp & 76.39±6.74 & 97.55±1.58 & 99.79±0.24 & 98.76±0.92 & 22.17±2.23 & 99.52±0.54 & 24.1±4.08 & 98.94±0.89 & 11.53±2.05 \\ 
 GBC & wisconsin & 70.47±12.32 & 98.35±1.73 & 98.86±1.7 & 99.05±1.53 & 19.77±2.19 & 98.66±1.81 & 14.47±2.05 & 98.02±2.02 & 4.63±1.4 \\ 
 GBC & SPECT & 29.82±20.71 & 73.32±41.32 & 73.57±41.44 & 73.72±41.54 & 22.27±2.23 & 73.52±41.43 & 4.03±2.55 & 73.57±41.44 & 1.23±0.77 \\ 
 GBC & salud-covid & 55.16±7.27 & 66.54±17.66 & 74.48±12.72 & 73.44±10.42 & 17.93±2.36 & 70.89±14.33 & 32.77±17.88 & 75.59±12.09 & 11.33±1.9 \\ 
 GBC & tic-tac-toe & 45.97±9.39 & 96.45±3.63 & 99.93±0.37 & 98.1±2.64 & 21.8±2.28 & 99.72±0.66 & 22.9±6.59 & 98.84±1.52 & 22.07±4.53 \\  \hline
 GBC & Total Average & 61.27±23.74 & 86.72±25.16 & 89.22±24.16 & 88.63±23.93 & 21.49±3.31 & 88.51±24.66 & 17.28±13.07 & 88.65±23.74 & 8.92±7.46 \\  \hline

 ADA & divorce & 90.98±16.33 & 87.51±29.9 & 87.83±29.94 & 87.8±29.98 & 24.87±3.55 & 86.82±29.64 & 3.53±0.51 & 86.42±29.59 & 3.17±0.7 \\ 
 ADA & kr-vs-kp & 62.01±12.06 & 95.31±1.51 & 98.15±0.98 & 97.7±1.07 & 21.03±3.46 & 48.56±9.55 & 2±0.0 & 98.45±1.0 & 73.77±9.45 \\ 
 ADA & wisconsin & 75.75±6.43 & 97.86±1.17 & 98.65±1.33 & 99.07±1.0 & 19.33±2.32 & 96.39±3.06 & 5.43±1.72 & 97.78±1.48 & 7.37±1.94 \\ 
 ADA & SPECT & 17.89±11.32 & 90.12±24.99 & 91.14±25.08 & 90.61±25.06 & 22±2.41 & 90.97±25.1 & 2±0.0 & 90.97±25.1 & 1.63±0.67 \\ 
 ADA & salud-covid & 38.72±9.32 & 37.0±33.33 & 39.9±33.97 & 32.65±30.5 & 18.57±2.21 & 8.06±19.44 & 2.87±1.72 & 18.37±30.46 & 10.4±2.93 \\ 
 ADA & tic-tac-toe & 54.08±7.91 & 95.66±4.18 & 99.6±0.8 & 97.18±2.96 & 21.1±2.07 & 81.03±3.12 & 1±0.0 & 99.08±1.21 & 20±1.84 \\ \hline
 ADA & Total Average & 56.57±26.24 & 83.91±29.74 & 85.88±29.63 & 84.17±30.82 & 21.15±3.38 & 68.64±36.05 & 2.81±1.73 & 81.84±35.02 & 19.39±25.45 \\ \hline

 CAT & divorce & 91.18±14.39 & 92.74±18.44 & 63.17±12.76 & 95.18±18.26 & 24.7±2.95 & 93.71±18.3 & 4.47±0.73 & 94.63±18.37 & 2.93±0.45 \\ 
 CAT & kr-vs-kp & 49.27±11.0 & 99.49±1.2 & 60.86±7.39 & 99.5±1.19 & 21.8±2.98 & 93.11±4.12 & 3.63±0.49 & 99.49±1.2 & 3.23±0.9 \\ 
 CAT & wisconsin & 84.23±4.53 & 98.74±1.27 & 82.9±2.1 & 99.25±1.04 & 19.2±2.62 & 97.05±3.03 & 5.87±1.66 & 99.19±0.87 & 8.23±1.41 \\ 
 CAT & SPECT & 33.8±18.6 & 85.72±29.34 & 86.39±29.5 & 86.53±29.54 & 22.3±2.48 & 86.39±29.5 & 2±0.0 & 86.39±29.5 & 2.3±1.18 \\ 
 CAT & salud-covid & 46.57±8.81 & 73.58±16.66 & 12.75±4.13 & 78.54±16.35 & 18.43±2.28 & 18.65±25.96 & 2.27±1.55 & 78.41±16.71 & 6.5±1.46 \\ 
 CAT & tic-tac-toe & 48.71±6.08 & 99.46±0.98 & 88.94±2.77 & 99.92±0.31 & 21±1.74 & 88.94±2.77 & 1±0.0 & 99.96±0.23 & 12.27±3.67 \\  \hline
 CAT & Total Average & 58.96±24.03 & 91.62±18.16 & 65.84±29.46 & 93.15±17.42 & 21.24±3.25 & 79.64±32.71 & 3.21±1.91 & 93.01±17.5 & 5.91±3.98 \\  \hline

        \textbf{Global Average} & Total Average & 60.17±24.31 & 88.16±23.56 & 82.31±28.53 & 89.38±23.45 & 21.36±3.42 & 81.89±30.21 & 10.43±10.76 & 88.76±24.91 & 10.82±14.69 \\ \hline
 \end{tabular}
 } 
 \label{table:results_table_non_recursive}
\end{table}

The performance of the non-recursive and recursive SRules models outperforms always to RuleCOSI (global average improvement of $\approx 7\%$) and maintains the F1 results obtained with RuleFIT but with a smaller number of rules. We have used a two-sample t-test for equal means to test the hypothesis that the obtained average classification rate and number of rules for the different models are not due to chance. The obtained overall significance values are shown in tables~\ref{results_table_no_recursive_global_pvalue}, ~\ref{results_table_recursive_global_pvalue} and by experiment in tables~\ref{table:results_table_non_recursive_pvalue}, \ref{table:results_table_recursive_pvalue} . We applied this test under the null hypothesis that the means are equal and the observations have different standard deviations. We conclude that the improvement obtained for the analysis SR vs. RC is significant for F1 metric (rejecting the hypothesis of equal means with a $p-value < 0.0001$) and similar for the number of rules ($p-value = 0.5640$); for the analysis of SR vs. RFIT the results are similar for F1 metric ($p-value = 0.6222$) and the improvement is significant for the number of rules ($p-value < 0.0001$); and for the recursive SRules the analysis RSR vs. RC improvement is significant for both F1 metric ($p-value <0.0001$) and number of rules ($p-value < 0.0001$); for the analysis of RSR vs. RFIT the results are similar for F1 metric ($p-value = 0.3611$) and the improvement is significant for the number of rules ($p-value < 0.0001$). Using the overall results we can conclude that the SRules maintains the results or improves it versus the other methods, and that recursive SRules improves always but for RFIT F1 metric than maintains similar results. 

If we are interested in high interpretability, we can observed that using Cat Boost model and the non-recursive SRules algorithm we obtain a 93.01\% F1-score with only 5.9 rules, improving on average the other methods in both criteria. This demonstrates the capabilities of the algorithm to achieve high interpretability and high performance sacrificing coverage. In contrast, there is a difference on coverage between non-recursive and recursive approaches. The difference is of 12.40\%, that is, non-recursive approach covers on average $\approx 60.2\%$ of the instances and the recursive $\approx 72.6\%$. Overall, the recursive approach maintains the classification performance  and increases the coverage. As a counterpart, the number of rules increase on average from 10.82 to 16.68 (as expected) sacrificing interpretability for coverage. The parameters search has been optimized using the F1-score x coverage criteria and table~\ref{table:metricF1Cov} shows the average values of this metric for each model and the non-recursive (SRules) and recursive approaches (RSRules). It can be seen that the best performance is obtained by Random Forest and CatBoost respectively. On the other hand, AdaBoost shows a clear decrease in performance over the other.

Analyzing the results for each of the dataset independently (see Tables~\ref{table:results_table_non_recursive} and ~\ref{table:results_table_recursive}) can be seen that most of the dataset and models benefit of the recursive approach and some are very significantly (e.g. kr-vs-kp  or SPECT ). In SPECT dataset, comparing CAT + NR-Srules with CAT + R-SRules we obtain an improvement on F1-score (86.39±29.5\% vs 95.87±4.16\%), an improvement in coverage (33.8±18.6\% vs 48.26±11.23\%), and a small increment in the number of rules (2.3±1.18 vs. 3.97±1.3). Furthermore,  if we compare these results with RuleFit or RuleCOSI our method significantly improves both in F1-score and in number of rules metrics (significance $p-values$ are shown in tables~\ref{table:results_table_non_recursive_pvalue}, \ref{table:results_table_recursive_pvalue} . Similarly,  comparing  CAT + SRules with CAT + RSRules for kr-vs-kp dataset we obtain similar performance (99.49±1.2\% vs. 96.76±3.21\%), a higher number of rules  (3.23±0.9 vs.  15.97±3.61), and a significantly improvement in coverage ( 49.27±11.0\% vs. 89.67±4.9\%). This is an example of the benefits of the proposed method, the non-recursive one may provide a simple interpretation via a small number of rules but covers only some instances of the dataset, and the recursive provides a better coverage (without sacrificing performance) but with a larger number of rules and, thereof, worsens the interpretability. The parameters of the method allows to adapt between these different scenarios according to the users' needs.

\subsection{Brief disccusion}
Note that all the results presented in this section are associated with the indicated coverage. That is, SRules is able to reduce coverage of the generated rules $R^*$ to maintain classification performance and interpretability, and hence the results are associated to that coverage as indicated for each experiment. It is also important note that among the three characteristics that we are trying to pursue in our work, high performance, high interpretation, and large coverage; high performance is the only one that we always try to maintain (both by the used optimization of the metric: F1-score x coverage, and by the learning process of the ensemble model that minimizes the classification error), and coverage and ease of interpretation are linked in such a way that an increase in one usually decreases the other. Thus, we try to explain with the generated rulesets for the areas of the learner model where the classification rate is high. This is an important difference with respect to other methods that rely entirely on the learned ensemble model and its structures to generate the ruleset, but without any control over the ruleset generation process. On the other hand, we can make decisions about which parts of the solution, learned by the models, we want to explain and focus only on those parts that are close to a perfect fit.   

\begin{table}
 \centering
 \caption{Recursive SRules (RSR). Comparison p-value performance per model and dataset.}

\resizebox{\columnwidth}{!}{%
 \begin{tabular}{c||c||c||c|c|cc|cc}
 \hline
       \textbf{Method} & \textbf{Dataset} & \textbf{Coverage}& \textbf{DT F1 p-value} & \textbf{Ensemble  F1 p-value}& \textbf{RFIT F1 p-value} & \textbf{RFIT  rl. num. p-value} & \textbf{RC  F1 p-value} & \textbf{RC  rl. num. p-value}  \\ \hline 

 RF & divorce & 95.69±4.87 & 0.4743 & 0.5361 & 0.8769 & <0.0001 & 0.5793 & <0.0001 \\ 
 RF & kr-vs-kp & 93.43±2.59 & 0.0615 & 0.0832 & 0.2681 & <0.0001 & 0.1698 & <0.0001 \\ 
 RF & wisconsin & 83.55±7.38 & 0.4913 & 0.1675 & 0.1664 & <0.0001 & 0.8116 & <0.0001 \\ 
 RF & SPECT & 39.53±16.74 & 0.8572 & 1.0000 & 0.9621 & <0.0001 & 0.8675 & <0.0001 \\ 
 RF & salud-covid & 60.3±5.42 & 0.1594 & 0.0002 & 0.8317 & <0.0001 & 0.5851 & <0.0001 \\ 
 RF & tic-tac-toe & 66.6±6.8 & 0.0039 & <0.0001 & 0.0128 & <0.0001 & 0.0019 & 0.0043 \\  \hline
 RF & Total Average & 73.18±21.63 & 0.2708 & 0.0373 & 0.9170 & <0.0001 & 0.8044 & <0.0001 \\  \hline

 GBC & divorce & 95.49±11.0 & 0.5451 & 0.4997 & 0.4448 & <0.0001 & 0.6057 & <0.0001 \\ 
 GBC & kr-vs-kp & 93.58±3.05 & 0.0006 & <0.0001 & <0.0001 & 0.0002 & <0.0001 & 0.002 \\ 
 GBC & wisconsin & 86.78±4.92 & 0.825 & 0.1499 & 0.0413 & <0.0001 & 0.3131 & <0.0001 \\ 
 GBC & SPECT & 34.62±23.48 & 0.9333 & 1.0000 & 0.9479 & <0.0001 & 1.0000 & 0.0393 \\ 
 GBC & salud-covid & 60.55±6.82 & 0.0093 & 0.6787 & 0.7563 & 0.0022 & 0.1347 & <0.0001 \\ 
 GBC & tic-tac-toe & 70.05±8.0 & 0.0042 & <0.0001 & 0.1380 & <0.0001 & <0.0001 & <0.0001 \\  \hline
 GBC & Total Average & 73.51±24.37 & 0.6104 & 0.2606 & 0.5779 & <0.0001 & 0.6666 & 0.0078 \\  \hline

 ADA & divorce & 94.71±6.8 & 0.2933 & 0.4542 & 0.0751 & <0.0001 & 0.9835 & 0.0743 \\ 
 ADA & kr-vs-kp & 84.74±6.83 & 0.5241 & 0.0042 & 0.0063 & <0.0001 & <0.0001 & <0.0001 \\ 
 ADA & wisconsin & 82.39±7.54 & 0.2560 & 0.0008 & 0.0034 & <0.0001 & 0.0007 & <0.0001 \\ 
 ADA & SPECT & 30.15±15.5 & 0.2849 & 0.8940 & 0.9340 & <0.0001 & 1.0000 & <0.0001 \\ 
 ADA & salud-covid & 53.82±5.62 & 0.0833 & 0.0221 & 0.1722 & 0.0103 & <0.0001 & <0.0001 \\ 
 ADA & tic-tac-toe & 73.74±7.22 & <0.0001 & 0.0014 & 0.0175 & <0.0001 & <0.0001 & <0.0001 \\ \hline
 ADA & Total Average & 69.92±23.53 & 0.7090 & 0.0945 & 0.3032 & 0.0117 & <0.0001 & <0.0001 \\ \hline

 CAT & divorce & 94.71±5.43 & 0.5821 & <0.0001 & 0.1352 & <0.0001 & 0.8598 & <0.0001 \\ 
 CAT & kr-vs-kp & 89.67±4.9 & 0.8069 & <0.0001 & 0.0256 & <0.0001 & <0.0001 & <0.0001 \\ 
 CAT & wisconsin & 87.95±5.42 & 0.0094 & <0.0001 & 0.8052 & <0.0001 & <0.0001 & <0.0001 \\ 
 CAT & SPECT & 48.26±11.23 & 0.3278 & 1.0000 & 0.7006 & <0.0001 & 1.0000 & <0.0001 \\ 
 CAT & salud-covid & 57.38±6.91 & 0.0935 & <0.0001 & 0.8129 & <0.0001 & <0.0001 & <0.0001 \\ 
 CAT & tic-tac-toe & 63.88±7.34 & 0.0005 & <0.0001 & 0.0214 & 0.8012 & <0.0001 & <0.0001 \\ \hline
 CAT & Total Average & 73.64±19.24 & 0.1610 & <0.0001 & 0.8750 & <0.0001 & <0.0001 & <0.0001 \\ \hline

 \textbf{Global Average}  & Total Average & 72.57±22.29 & 0.2651 & <0.0001 & 0.3611 & <0.0001 & <0.0001 & <0.0001 \\ \hline

 \end{tabular}
 } 
 \label{table:results_table_recursive_pvalue}
\end{table}
\begin{table}
 \centering
 \caption{Recursive SRules (RSR). Average performance per model and dataset.}

\resizebox{\columnwidth}{!}{%
 \begin{tabular}{c||c||c||c|c|cc|cc|cc}
 \hline

\textbf{Method}& \textbf{Dataset}  & \textbf{Coverage} & \textbf{DT F1} & \textbf{Ensemble F1} & \textbf{RFIT F1} & \textbf{RFIT rl. num.} & \textbf{RC F1} & \textbf{RC rl. num.} & \textbf{RSR F1} & \textbf{RSR rl. num.} \\ \hline

 RF & divorce & 95.69±4.87 & 96.72±4.42 & 98.11±3.75 & 97.66±4.24 & 25.53±3.32 & 96.91±4.25 & 5.53±1.17 & 97.49±3.9 & 3.33±0.84 \\ 
 RF & kr-vs-kp & 93.43±2.59 & 95.97±1.66 & 96.1±1.45 & 98.05±1.08 & 21.93±3.77 & 98.24±1.2 & 16.23±4.01 & 97.31±3.46 & 29.5±4.22 \\ 
 RF & wisconsin & 83.55±7.38 & 97.59±1.9 & 98.44±1.45 & 98.43±1.41 & 19.57±2.65 & 97.78±2.06 & 16.73±3.49 & 97.9±1.56 & 7.5±1.83 \\ 
 RF & SPECT & 39.53±16.74 & 88.18±24.74 & 89.33±24.64 & 89.03±24.6 & 22.47±2.08 & 88.27±24.57 & 20.53±2.64 & 89.33±24.64 & 2.27±1.53 \\ 
 RF & salud-covid & 60.3±5.42 & 60.31±20.66 & 43.07±26.87 & 66.85±20.86 & 17.97±2.82 & 65.03±20.85 & 26.43±10.1 & 68.01±21.15 & 13.17±2.97 \\ 
 RF & tic-tac-toe & 66.6±6.8 & 97.31±1.7 & 95.43±2.39 & 97.49±1.72 & 20.97±2.17 & 99.67±0.66 & 24.7±4.24 & 98.61±1.66 & 27.87±4.02 \\ \hline
 RF & Total Average & 73.18±21.63 & 89.35±18.78 & 86.75±24.73 & 91.25±17.4 & 21.41±3.69 & 90.98±17.93 & 18.36±8.52 & 91.44±17.21 & 13.94±11.39 \\ \hline
 GBC & divorce & 95.49±11.0 & 93.67±18.15 & 94.01±18.17 & 94.42±18.08 & 25.6±3.01 & 93.24±18.01 & 5.43±0.63 & 90.81±18.26 & 2.83±0.59 \\ 
 GBC & kr-vs-kp & 93.58±3.05 & 96.26±1.15 & 99.76±0.3 & 98.13±1.09 & 22.67±2.8 & 99.36±0.59 & 23±4.76 & 93.95±3.3 & 19.7±2.93 \\ 
 GBC & wisconsin & 86.78±4.92 & 97.09±2.08 & 97.85±2.28 & 98.12±1.78 & 19.87±2.24 & 97.55±2.0 & 14.8±3.7 & 96.96±2.47 & 7.87±1.28 \\ 
 GBC & SPECT & 34.62±23.48 & 81.77±32.88 & 82.49±33.13 & 81.93±33.03 & 22.13±2.62 & 82.49±33.13 & 4.33±3.39 & 82.49±33.13 & 2.77±2.25 \\ 
 GBC & salud-covid & 60.55±6.82 & 65.42±12.8 & 75.64±10.62 & 73.35±12.4 & 17.83±1.51 & 69.29±12.98 & 33.47±19.98 & 74.36±12.92 & 16.03±2.68 \\ 
 GBC & tic-tac-toe & 70.05±8.0 & 96.2±2.64 & 99.86±0.35 & 97.11±2.5 & 20.8±2.07 & 99.77±0.48 & 22.07±5.23 & 97.99±1.96 & 33.03±3.9 \\  \hline
 GBC & Total Average & 73.51±24.37 & 88.4±19.77 & 91.6±18.36 & 90.51±18.64 & 21.48±3.41 & 90.29±19.54 & 17.18±13.52 & 89.43±18.27 & 13.71±11.01 \\  \hline

 ADA & divorce & 94.71±6.8 & 95.2±4.62 & 97.2±3.68 & 98.23±3.42 & 25.53±2.86 & 96.44±4.28 & 3.63±0.49 & 96.42±4.29 & 3.3±0.88 \\ 
 ADA & kr-vs-kp & 84.74±6.83 & 95.51±1.23 & 97.86±0.94 & 97.74±1.24 & 22.1±2.29 & 49.94±4.29 & 2±0.0 & 94.86±5.42 & 100.5±17.13 \\ 
 ADA & wisconsin & 82.39±7.54 & 97.68±1.94 & 98.6±1.41 & 98.57±1.87 & 19.3±2.49 & 94.56±3.48 & 5.23±1.76 & 97.12±1.79 & 9.47±2.18 \\ 
 ADA & SPECT & 30.15±15.5 & 93.95±7.5 & 96.28±7.46 & 96.18±7.4 & 22.2±2.41 & 96.02±7.4 & 2±0.0 & 96.02±7.4 & 3.73±1.62 \\ 
 ADA & salud-covid & 53.82±5.62 & 55.41±22.54 & 59.28±23.61 & 53.59±25.16 & 18±2.03 & 18.0±22.93 & 2.6±1.3 & 44.81±24.06 & 20.83±5.49 \\ 
 ADA & tic-tac-toe & 73.74±7.22 & 94.65±3.21 & 99.62±0.65 & 96.94±2.69 & 21.73±2.05 & 78.87±2.75 & 1±0.0 & 98.41±1.86 & 30.53±3.41 \\  \hline
 ADA & Total Average & 69.92±23.53 & 88.73±17.94 & 91.47±17.65 & 90.21±19.64 & 21.48±3.35 & 72.31±31.04 & 2.74±1.64 & 87.94±22.06 & 28.06±34.71 \\  \hline

 CAT & divorce & 94.71±5.43 & 95.48±5.65 & 66.05±3.89 & 97.91±3.43 & 26.37±3.9 & 96.05±3.42 & 4.73±0.94 & 96.24±4.96 & 3.53±0.78 \\ 
 CAT & kr-vs-kp & 89.67±4.9 & 96.92±1.41 & 71.23±2.5 & 98.18±1.13 & 22.1±3.35 & 91.46±4.05 & 3.73±0.45 & 96.76±3.21 & 15.97±3.61 \\ 
 CAT & wisconsin & 87.95±5.42 & 97.95±1.59 & 81.72±1.93 & 99.0±1.11 & 18.83±2.49 & 96.74±2.42 & 6.03±1.97 & 98.92±1.19 & 9.33±1.69 \\ 
 CAT & SPECT & 48.26±11.23 & 94.69±5.04 & 95.87±4.16 & 95.48±3.72 & 22.53±2.33 & 95.87±4.16 & 2±0.0 & 95.87±4.16 & 3.97±1.3 \\ 
 CAT & salud-covid & 57.38±6.91 & 66.43±16.27 & 16.24±4.85 & 72.25±14.98 & 17.97±2.24 & 10.68±20.78 & 2±1.05 & 73.14±14.13 & 11.53±2.29 \\ 
 CAT & tic-tac-toe & 63.88±7.34 & 97.28±2.08 & 84.52±3.37 & 98.04±1.51 & 21.6±1.79 & 84.52±3.37 & 1±0.0 & 98.78±0.85 & 21.8±3.94 \\ \hline
 CAT & Total Average & 73.64±19.24 & 91.46±13.45 & 69.27±25.88 & 93.48±11.54 & 21.57±3.87 & 79.22±32.29 & 3.25±2.02 & 93.29±11.14 & 11.02±6.94 \\ \hline

     \textbf{Global} & Total Average & 72.57±22.29 & 89.48±17.66 & 84.77±23.76 & 91.36±17.11 & 21.48±3.58 & 83.2±27.13 & 10.38±10.96 & 90.52±17.69 & 16.68±20.47 \\  \hline
 \end{tabular}
 } 
 \label{table:results_table_recursive}
\end{table}

\begin{table}
    \centering
    \caption{Average 'F1-score x coverage' and number of rules per model}

\resizebox{\columnwidth}{!}{%
    \begin{tabular}{|c|||c|c|c|c|}
    \hline
        \textbf{Method} & \textbf{SRules rl. num.} & \textbf{SRules F1 x coverage} & \textbf{RSRules rl. num.}  & \textbf{RSRules F1 x coverage}\\ \hline
        Random Forest & 9.05±6.58 & 58.48±4.07 &   13.94±11.39 &  66.92±3.72 \\ 
        Gradient Boost & 8.92±7.46 & 54.31±5.64 &  13.71±11.01 &  65.74±4.45 \\ 
        Ada Boost & 19.39±25.45 & 46.3±9.19 &  28.06±34.71 &  61.49±5.19 \\ 
        Cat Boost & 5.91±3.98 & 54.84±4.2 &  11.02±6.94 &  68.7±2.14 \\   \hline

\end{tabular}
    }    
    \label{table:metricF1Cov}
\end{table}

\section{Conclusions}
In this paper, we proposed SRules, a method for providing interpretability to black-box machine learning models using recursively surrogate binary tree models. The method uses the most important features to create a binary decision tree containing a set of possible important rules. A conditional chi-square analysis selects the rules that have sufficient significance to be considered valid according to the imposed requirement of interpretability of the application. A final phase of pruning detects duplicates and combines all the rulesets obtained for the different areas of the learned model. The combination is made on the basis of the generality of the rules; the most general rules are retained, avoiding duplication and  improving interpretability.  In addition, several experiments were performed to evaluate and validate the obtained results, and it was shown that, in general, the performance of the obtained rulesets is maintained, whereas the number of rules is reduced. The main contribution of the proposed method relies on the ability to leverage between coverage and interpretability without losing performance, which is a great advantage for practitioners and is meaningful for understanding the main causes in the model decision. This is specially useful in domains such as medical diagnosis or biology. 
Our work has some limitations, such as the lack of support for continuous inputs because the algorithm is designed for binary surrogate decision trees or the comparison with other well known model agnostic feature importance techniques such as SHAP or LIME to extend its use to machine learning models that doesn't have intrinsic feature importance extraction methods. As future work we intend to include other machine learning models such as Neural Networks using model agnostic feature importance extraction techniques and add the functionalities to work with continuous features. 

\begin{acks}
This work has been funded by the project "Inteligencia Artificial eXplicable" IAX grant of the Young Researchers 2022/2024 initiative of the Community of Madrid, Spain. 
\end{acks}

\bibliographystyle{ACM-Reference-Format}
\bibliography{mybibliography}
\appendix









\end{document}